\pdfoutput=1

\documentclass[11pt]{article}

\usepackage[hyphens]{url}

\usepackage[final]{acl}

\usepackage{times}
\usepackage{latexsym}

\usepackage[T1]{fontenc}

\usepackage[utf8]{inputenc}

\usepackage{microtype}

\usepackage{inconsolata}

\usepackage{graphicx}

\usepackage{soul}
\renewcommand\hl[1]{#1}
 
\title{Attacking Misinformation Detection Using Adversarial Examples Generated by Language Models}

\author{\textbf{Piotr Przyby{\l}a}\textsuperscript{\textnormal{1,2}} \and 
        \textbf{Euan McGill}\textsuperscript{\textnormal{1}} \and
        \textbf{Horacio Saggion}\textsuperscript{\textnormal{1}}\\
  \textsuperscript{\textnormal{1}} TALN Group, Universitat Pompeu Fabra, Barcelona, Spain \\
  \textsuperscript{\textnormal{2}} Institute of Computer Science, Polish Academy of Sciences, Warsaw, Poland  \\
  {\tt \{piotr.przybyla, euan.mcgill, horacio.saggion\}@upf.edu}
 }

\begin{document}
\maketitle
\begin{abstract}
Large language models have many beneficial applications, but can they also be used to attack content-filtering algorithms in social media platforms? We investigate the challenge of generating adversarial examples to test the robustness of text classification algorithms detecting low-credibility content, including propaganda, false claims, rumours and hyperpartisan news. We focus on simulation of content moderation by setting realistic limits on the number of queries an attacker is allowed to attempt. Within our solution (TREPAT), initial rephrasings are generated by large language models with prompts inspired by meaning-preserving NLP tasks, such as text simplification and style transfer. Subsequently, these modifications are decomposed into small changes, applied through beam search procedure, until the victim classifier changes its decision. \hl{We perform (1) quantitative evaluation using various prompts, models and query limits, (2) targeted manual assessment of the generated text and (3) qualitative linguistic analysis}. The results confirm the superiority of our approach in the constrained scenario, especially in case of long input text (news articles), where exhaustive search is not feasible.
\end{abstract}

\section{Introduction}

Modern machine learning (ML) methods have proven effective in determining credibility of text in various scenarios \citep{Horne2017,Graves2018,Al-Sarem2019,Martino2020a,10.1007/978-3-031-56069-9_62}, helping to tackle the challenge of misinformation \citep{Lewandowsky2017,Tucker2018}. Because of this development, many large platforms hosting user-generated data, e.g. social media, use text classifiers as part of their content moderation systems \citep{Singhal2022}. This raises the need to assess the \textit{robustness} of such solutions, i.e. their ability to deliver correct result even for input manipulated by malicious actors. This is performed by seeking \textit{adversarial examples} (AEs) -- text samples modified in such a way that preserves their meaning, but elicits an incorrect response from the victim classifier \citep{Carter2021}.

A variety of experiments have been performed to confirm the vulnerability of credibility assessment to generic AE generation methods, and then seeking solutions tuned for this specific scenario within the \textit{InCrediblAE} shared task \citep{clef-checkthat:2024:task6}. The best approaches are based on iterative replacement of individual words with equivalents suggested by a language model. Generally, this direction has two major weaknesses.

Firstly, covering the vast space of possible rephrasing requires sending many queries to the victim system, sometimes several thousand, just to generate one adversarial AE. This makes the experiment stray far from a real-world implementation scenarios, where an adversary would be blocked from using the system when attempting to send so many queries. Secondly, word-replacement strategy can lead to poor meaning preservation. The manual evaluation of the shared task indicated that these methods often modify the meaning of the whole phrase, making such an AEs unusable.

However, AE generation is not the only task in Natural Language Processing (NLP) that requires modifying a given text while preserving its meaning, cf. text simplification \citep{Shardlow2014}, style transfer \citep{Pang2019} or paraphrasing \citep{zhou-bhat-2021-paraphrase}. The approaches using generative Large Language Models (LLMs) with carefully crafted prompts \citep{Jayawardena_2024,Kew2023,mukherjee-etal-2024-large-language} achieve the best results in these tasks.

Inspired by this work, here we propose TREPAT (\textbf{T}racing \textbf{RE}cursive \textbf{P}araphrasing for \textbf{A}dversarial examples from \textbf{T}ransformers): a solution for generating adversarial examples in English that leverages the LLMs' ability to reformulate a given text. The variants generated by LLMs are decomposed into atomic changes using Wagner-Fischer algorithm \citep{10.1145/321796.321811}, which are then recursively applied using beam search \citep{Lowerre1976}, until the victim classifier changes its decision.

The contributions of this article are as follows:
{
\setlength{\parskip}{-0.5em}
\begin{enumerate}
\setlength\itemsep{-0.2em}
    \item A novel method for employing generative LLMs to obtain numerous variants of a given text fragment that could serve as AEs,
    \item An investigation on which models and rephrasing prompts (focused on simplification, style change, paraphrasing etc.) return variants that change the victim's decision,
    \item Manual annotation of the examples generated by various methods to verify their meaning preservation and language naturalness,
    \item A linguistic analysis of the changes that LLMs perform in this scenario.
\end{enumerate}
}
The code for TREPAT and annotation results are openly shared to encourage further research\footnote{\url{https://github.com/piotrmp/trepat}}.

\section{Related work}

The investigation of AEs was initially proposed for image classification \citep{Szegedy2013} and the extension of the framework to the text domain is challenging due to discrete nature of the medium \citep{Zhang2020b}. In the misinformation domain, the fact-checking task was the first to be investigated for robustness \citep{Hidey2020,Zhou2019b}, followed by fake news detection \citep{Ali2021a,Koenders2021}.

A systematic analysis of AEs in credibility assessment, covering various tasks, attackers and victims, was performed through the BODEGA framework \citep{verifying_robustness}, highlighting the vulnerabilities that affects also very large models. This line was extended through the \textit{InCrediblAE} shared task \citep{clef-checkthat:2024:task6} organised at \textit{CheckThat!} evaluation lab \cite{clef-checkthat:2024-lncs}. The submitted solutions can be broadly divided into those relying on character changes (e.g. swapping \textit{0} for \textit{O}) \citep{clef-checkthat:2024:sinai,clef-checkthat:2024:turquaz,clef-checkthat:2024:texttrojaners}, replacing words according to the candidates from language models \citep{clef-checkthat:2024:palori,clef-checkthat:2024:openfact}, or both \citep{clef-checkthat:2024:mmu}. We can also mention XARELLO \citep{przybyla-etal-2024-know}, which is using the BODEGA data, but with a different usage scenario: assuming that the attacker is performing multiple attacks on the same victim in the \textit{adaptation} phase and thus can learn what modifications are successful. This process, powered by reinforcement learning, can allow to greatly reduce the number of queries in the test phase, but might not be possible if the victim is updated frequently. Additionally, the robustness analysis has also been performed to test attacks on the task of machine-generated text detection \cite{wang-etal-2024-raft}.

Only one of the methods at InCrediblAE used an LLM model to generate rephrasings and the results were not satisfying \citep{clef-checkthat:2024:turquaz}. Even beyond the misinformation detection, the abilities of LLMs have not yet been fully utilised for AE generation. We can mention their use to perform two subtasks: word importance ranking and synonym generation \cite{10580402}. Moreover, \textit{PromptAttack} \citep{DBLP:conf/iclr/XuKLC0ZK24} involves prompting an LLM to generate AEs, which makes it similar to our work. However, there is an important difference: PromptAttack assumes that the AEs are produced through interaction with the same model that is its victim. This approach cannot be applied to content filtering, where the model is inaccessible and might not even be a generative LLM.

\hl{Beyond the search for adversarial examples, there are  other ways to 'attack' LLMs, e.g. looking for prompts that make them produce a desired output (e.g. toxic text)} \citep{wallace-etal-2019-universal,10.1145/3689217.3690621}. However, there have been no successful approaches to use LLMs to generate reformulations that could be used as AEs to attack credibility assessment systems. This is the aim of TREPAT.

\section{Methods}

\begin{figure*}
\centering
\includegraphics[width=\linewidth]{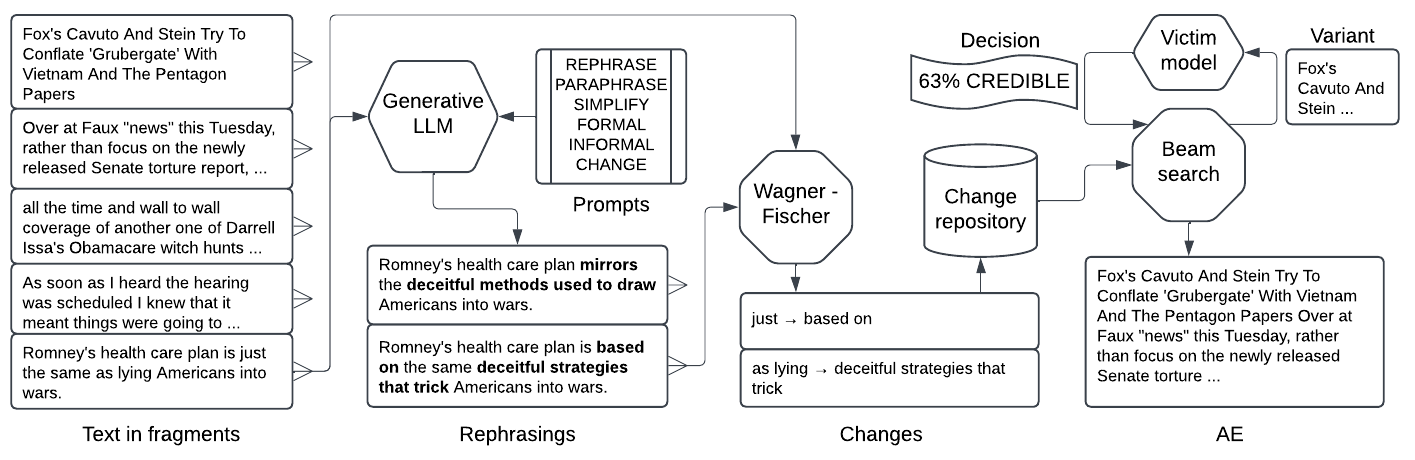} 
\caption{The architecture of TREPAT. A continuous text document is divided into fragments, each rephrased by a generative LLM (further processing only for one fragment is shown in the figure). The comparison of a rephrasing and text original text yields changes stored in a repository. These changes are iteratively applied following a beam search algorithm guided by the victim's response (i.e. binary credibility label).}
\label{fig:architecture}
\end{figure*}

TREPAT explores adversarial examples in several simple steps (see Figure \ref{fig:architecture}). Firstly, the text is split into smaller \textit{fragments} (Section \ref{sec:splitting}). Each of these fragments is then fed into an LLM with one of various prompts to obtain \textit{rephrasings} (Section \ref{sec:rephrasing}). Then, each rephrasing is decomposed into individual \textit{changes}, which are then subsequently applied to the original text, resulting in \textit{variants}, using the beam search procedure guided by the credibility score returned by the victim model (Section \ref{sec:changes}). Finally, if no AE has been found after testing all changes, the process is re-initiated with the best candidate seen so far as the starting point.

\subsection{Splitting}
\label{sec:splitting}

Our preliminary tests have shown that LLMs, especially smaller ones, struggle to rephrase long sentences in a single round, often omitting important parts. Moreover, some of the input instances consist of whole documents containing multiple sentences, e.g. news articles (see section \ref{sec:evaluation}).

To divide input text into fragments fit for rephrasing, we perform the following splitting operations\footnote{These rules were established based on a manual analysis of the output of the rephrasing module (for development portion of PR dataset, GEMMA-2B model, BiLSTM and BERT victims). We noticed that fragments shorter than 60 characters lacked the sufficient context for the LLM to rephrase, resulting in loss of meaning. Moreover, punctuation proved to be a good opportunity to separate longer phrases while maintaining semantic consistency.}:
{
\setlength{\parskip}{-0.5em}
\begin{enumerate}
\setlength\itemsep{-0.2em}
    \item Splitting the input types --  this only applies to examples which combine the evidence and the claim in a single input (see section \ref{sec:evaluation}),
    \item Splitting on newline characters,
    \item Splitting into sentences with LAMBO \citep{Przybya2022},
    \item Splitting on characters indicating phrase boundaries: dashes, quotation marks, commas and colons; as long as the results are at least 60 characters long.
\end{enumerate}
}
For each of the fragments we also preserve its offset to be able to combine changes to different fragments in a single output text.

\subsection{Rephrasing}
\label{sec:rephrasing}

The goal of this stage is to rephrase a given fragment in a way that changes its appearance but preserves the meaning. We use a LLM and provide it with a prompt corresponding to one of six commands, inspired with other text modification tasks:
{
\setlength{\parskip}{-0.5em}
\begin{itemize}
\setlength\itemsep{-0.2em}
    \item \texttt{REPHRASE}: the basic prompt, asking the model to \textit{rephrase} the input fragment,
    \item \texttt{PARAPHRASE}: a variant aimed at stronger meaning preservation, asking for a \textit{paraphrase}. LLMs have been shown to produce high-quality and diverse paraphrases \citep{Jayawardena_2024}.
    \item \texttt{SIMPLIFY}: a prompt requesting the model provide a simpler equivalent of the fragment. Previous work indicates that LLMs are able to handle this task based on a short prompt (with a few examples) \citep{Kew2023}.
    \item \texttt{FORMAL} and \texttt{INFORMAL}: variants requesting the model to rewrite the text in more (or less) formal style. Evaluations have shown that LLMs can achieve good-quality style transfer, at least in English \citep{mukherjee-etal-2024-large-language}.
    \item \texttt{CHANGE}: a different phrasing, explicitly emphasising the need to \textit{make changes} and relaxing the meaning preservation condition (\textit{try to preserve \ldots}). In preliminary experiments this has led to more aggresive modifications.
\end{itemize}
}
The prompts were formulated through experimenting with GEMMA 1.0 2B model \citep{GemmaTeam2024}. All of the prompts are quoted in Appendix \ref{sec:prompts}. Note how these prompts focus on the meaning-preservation goals, rather than expected modifications (e.g. \textit{Replace at most two words in the sentence\ldots}) used in PromptAttack \citep{DBLP:conf/iclr/XuKLC0ZK24}.

We use six pre-trained instruction-tuned LLMs of various sizes, obtained through \textit{HuggingFace Transformers} \cite{wolf-etal-2020-transformers}:
{
\setlength{\parskip}{-0.5em}
\begin{itemize}
\setlength\itemsep{-0.2em}
    \item \texttt{LLAMA1B}: Llama 3.2\footnote{\url{https://github.com/meta-llama/llama-models/blob/main/models/llama3_2/MODEL_CARD.md}} with 1 billion parameters\\(\texttt{meta-llama/Llama-3.2-1B-Instruct})
    \item \texttt{GEMMA2B}: Gemma 2.0 \citep{gemma_2024} with 2 billion parameters\\(\texttt{google/gemma-2-2b-it}),
    \item \texttt{LLAMA3B}: Llama 3.2 with 3 billion parameters (\texttt{meta-llama/Llama-3.2-3B-Instruct}),
    \item \texttt{OLMO7B}: OLMo \citep{groeneveld-etal-2024-olmo} v. 0724  with 7 billion parameters\\(\texttt{allenai/OLMo-7B-0724-Instruct-hf}),
    \item \texttt{LLAMA8B}: Llama 3.1 with 8 billion parameters\\(\texttt{meta-llama/Llama-3.1-8B-Instruct}),
    \item \texttt{GEMMA9B}: Gemma 2.0 with 9 billion parameters (\texttt{google/gemma-2-9b-it}),
\end{itemize}
}
The output of an LLM is parsed by splitting it into newline-separated reformulations and trimming unnecessary elements (enumerations, end-of-text tokens etc.).

\subsection{Obtaining changes}
\label{sec:changes}

Our preliminary experiments have shown that the reformulations generated by LLMs are usually not directly useful as AEs. They contain numerous modifications, while good-quality AEs can differ from the original example by only a single word. This is why we decompose the obtained reformulations into individual \textit{changes}, each of which corresponds to a continuous sequence of tokens being replaced by a different sequence of tokens.

Take the following example:
{
\setlength{\parskip}{-0.5em}
\begin{itemize}
\setlength\itemsep{-0.2em}
    \item INPUT: \textit{The recent rise of food prices is resulting in widespread discontent.}
    \item LLM OUTPUT: \textit{The recent surge in food prices has caused widespread unease.}
\end{itemize}
}
This reformulation, performed by an LLM (GEMMA 2B), includes three changes\footnote{Note that changes can be context-sensitive, even if these examples appear general. Thus, in TREPAT we only apply changes to the sentences they were extracted from.}:
{
\setlength{\parskip}{-0.5em}
\begin{itemize}
\setlength\itemsep{-0.2em}
    \item \textit{rise of} -> \textit{surge in}
    \item \textit{is resulting in} -> \textit{has caused}
    \item \textit{discontent} -> \textit{unease}
\end{itemize}
}
The LLM has made multi-token changes, which would not be possible with methods based on word replacements, e.g. BERT-ATTACK.

In order to obtain these changes we convert both text fragments into sequences of tokens and then apply the Wagner-Fischer algorithm \citep{10.1145/321796.321811} for computing the edit distance. It represents a reformulation through the means of ADD, DELETE and REPLACE operations. We aggregate neighbouring operations to allow for multi-token changes, as shown above.

Finally, the changes are filtered by discarding those that contain only ADD or DELETE operations\footnote{These often correspond to reformulations where text content is moved within a fragment, which do not preserve meaning when decomposed into individual changes.} or modify more than 2/3 of the fragment or 1/3 of the whole text.

\subsection{Applying changes}
The changes obtained from all reformulations of all fragments are collected in a single repository in order to be applied to the input text. Then, text \textit{variants} are created by starting from the original text and gradually adding changes that modify it. Each created variant is sent as a query to the victim classifier and if it results in a modified response, it is returned as a successful AE.

In the example cited previously (\textit{The recent rise of food prices is resulting in widespread discontent.}), the algorithm can check the victim's response to variants that include just one of the changes, e.g. \textit{The recent \textbf{surge in} food prices is resulting in widespread discontent.} or \textit{The recent rise of food prices \textbf{has caused} widespread discontent.} Then, variants combining two changes are possible, e.g. \textit{The recent \textbf{surge in} food prices is resulting in widespread \textbf{unease}.}, with three changes, etc.

However, given the limitation of queries (see Section \ref{sec:evaluation}) and the number of possible changes in longer text examples, it is impossible to test all combinations to find the ones that change the victim's decision. Inspired by one of the solutions at the InCrediblAE shared task \citep{clef-checkthat:2024:texttrojaners}, we apply beam search \citep{Lowerre1976}. This means that we record the \textit{value} of each variant (i.e. the reduction of the probability of the original class according to victim classifier) and at any stage only $k$ variants with the highest value are kept for applying further changes. The $k$ is set to 5 to reduce the size of search scope\footnote{This beam size corresponds to the typical number of atomic changes obtained from a single rephrasing, observed in the preliminary experiments. Thus, if only one useful rephrasing is generated, all of its changes can be explored.}.

A change is only applicable to a variant if the part of the text it modifies has not been modified yet (by itself or another overlapping change). If at some point we run out of available changes, a new batch of reformulations is generated by the LLM from the variant of highest value so far.

\section{Evaluation}
\label{sec:evaluation}
The evaluation is performed using the BODEGA framework \citep{verifying_robustness}, created to verify the robustness of credibility assessment solutions and based on previous corpora for English \cite{potthast-etal-2018-stylometric,Martino2020a,Thorne2018,Han2019}. It covers four misinformation detection tasks: propaganda recognition (PR), fact-checking (FC), rumour detection (RD) and hyperpartisan news classification (HN). These tasks include text with various length and features: individual sentences (PR), claims with relevant evidence (FC), Twitter threads (RD) and news articles (HN). For each of these tasks, cast as binary classification, a model is trained using one of four popular architectures: BiLSTM neural network \citep{Liu2019} and fine-tuned BERT \citep{Devlin2018} or GEMMA \citep{GemmaTeam2024} in 2-billion and 7-billion variants. 

BODEGA evaluates a given adversarial example by comparing it to the original text and measuring \textit{confusion}, checking if the victim classifier changed its prediction; \textit{semantic} similarity between the two texts using BLEURT \citep{sellam-etal-2020-bleurt}; and \textit{character} similarity using Levenshtein distance \citep{1965}. All three scores are expressed as numbers in 0-1 range and are multiplied for a single BODEGA score, but can also be interpreted separately for better understanding of the results.

Additionally, we introduce a limit on the number of queries an attacker can perform to make our evaluation closer to real-world scenarios. The maximum numbers of posts that social media allow a user to submit are not disclosed, but estimated between 10 and 100 submissions per day\footnote{E.g. \url{https://help.simplified.com/en/articles/6067588-what-are-the-daily-posting-limits-on-each-social-media} or \url{https://support.buffer.com/article/646-daily-posting-limits}}. We decided to generally allow 50 attempts, but also test how this parameter influences the performance (see Experiment 4).

\subsection{Automatic experiments}

We perform four experiments:

\textbf{Experiment 1} aims to compare various LLMs (section \ref{sec:rephrasing}) in the task. We use the REPHRASE prompt, and for each task take 400 examples from the development portion of the BODEGA datasets and compute BODEGA score averaged over all victim models. The results of this experiment guide the choice of an LLM for next steps.

\textbf{Experiment 2} is designed to check which prompting strategy (section \ref{sec:rephrasing}) is the most effective for generating AEs that achieve decision change and preserve the meaning. It also involves development data and follows the design of experiment 1, except we only test the LLM chosen there. The prompt (or prompts) selected based on these results will be used in final evaluation.

\textbf{Experiment 3} plays the role of the main evaluation. It is based on the attack portion of the BODEGA datasets and compares TREPAT with parameters chosen as above with all the baselines. Unlike in the previous experiments, we analyse the results for each victim separately and include the partial scores -- for confusion, semantic and character similarity.

\textbf{Experiment 4} tests the applicability of the proposed method by checking the performance (BODEGA score averaged over victims and tasks) of TREPAT and baselines when different number of queries to a victim are allowed: 10, 50 (default used in experiments 1-3), 100 or 250.

\subsection{Baselines}

Based on the analysis of the previous work, the following solutions are used to compare to full TREPAT in experiments 3 and 4:

\textbf{BERT-ATTACK} \citep{li-etal-2020-bert-attack} was the overall best method in the original BODEGA evaluation, which covered a variety of AE generation approaches. It looks for replacements to a given word by applying language modelling through BERT. 

\textbf{F-BERT-ATTACK} is our modification of the above to better fit the constrained query limit. The problematic aspect of BERT-ATTACK is its initial step, which selects the most vulnerable word by observing victim's response to its removal, requiring sending many queries for longer text. Here we replace this step by obtaining word importance randomly, allowing the attacker to perform viable attacks from the first query.

\textbf{BeamAttack} is a solution submitted by the \textit{TextTrojaners} team  \citep{clef-checkthat:2024:texttrojaners} to the \textit{InCrediblAE} shared task, obtaining nearly the highest score. Similarly to our solution, it employs beam search to find the best replacement. We set its parameters to the lowest values considered by the authors to limit the search scope: 10 beams, 5 hypotheses and branching factor of 10.

\hl{\textbf{TREPAT-simple} is a simplified version of TREPAT, where LLM-generated rephrasings are used directly, instead of being split into changes and applied through beam search. The variant with the whole procedure is labelled \textbf{TREPAT-full}.}

\subsection{Manual evaluation}
\label{sec:manualevaluation}

Evaluating meaning similarity is a difficult task and while automatic measures are being used in an equivalent role in machine translation, in InCrediblAE they were shown to poorly align with the human judgement in the adversarial example assessment. For this reason, we have decided to perform manual evaluation by asking human annotators to evaluate the quality of AEs generated by TREPAT, compared to other methods.

For that purpose, we take the output of Experiment 3 and randomly select 20\% of the cases where successful AEs are available: from TREPAT and the best baseline method for this victim/task combination. We use two annotators, with random 25\% of the data being assigned to both of them to measure agreement. Both our annotators are linguists: one a native speaker of English and the other one with certified proficiency in the language.

Each annotator is presented with a list of triples, consisting of the original text, modification A and modification B, where A and B are randomly taken from the baseline or TREPAT adversarial examples. The spans changed between the variants are highlighted. The annotators are then asked to decide which modification offers better \textbf{meaning preservation} (maintaining the meaning expressed in the original text) and language \textbf{naturalness} (seeming fluent, grammatical and authentic, as opposed to artificial and manipulated). While the latter criterion is well known in human evaluation of NLP solutions \citep{howcroft-etal-2020-twenty,belz-etal-2020-disentangling}, the former is specific to AE assessment, but also found in evaluation of style transfer \citep{cao-etal-2020-expertise} and simplification \cite{stodden-kallmeyer-2022-ts}.

When the annotators are unable to choose between A and B, they can say that either 'Both' or 'Neither' of the options satisfies a given criterion. However, they are encouraged to make a clear decision even for small differences. Full annotation guidelines are included as Appendix \ref{sec:guidelines}.

\section{Results}

\hl{The work performed offers us three perspectives for assessing the quality of the generated examples:}
{
\setlength{\parskip}{-0.5em}
\begin{itemize}
\setlength\itemsep{-0.2em}
    \item \hl{automatic quantitative experiments, measuring the interaction between TREPAT and various attack victims} (\ref{sec:automaticresults}),
    \item \hl{manual evaluation of the text quality, comparing our method with the strongest competitor, done blindly by proficient speakers} (\ref{sec:manualresults}),
    \item \hl{qualitative analysis of selected examples by a professional linguist to investigate the linguistic patterns present in the generations} (\ref{sec:ling}).
\end{itemize}
}
\subsection{Automatic experiments}
\label{sec:automaticresults}
\begin{table}
    \small
    \begin{tabular}{r|rrrr}
    \hline
    & \multicolumn{4}{|c}{\textbf{BODEGA score}}\\
    \textbf{LLM} & \textbf{PR} & \textbf{FC} & \textbf{RD} & \textbf{HN} \\
    \hline
        LLAMA 1B  & {0.2297} & {0.3007} & \textbf{0.1377} & \textbf{0.1691} \\
        GEMMA 2B  & {0.2119} & {0.2316} & {0.0960} & {0.1512} \\
        LLAMA 3B  & {0.2231} & \textbf{0.3062} & {0.1188} & {0.1548} \\
        OLMO 7B  & {0.2420} & {0.3036} & {0.1313} & {0.1408} \\
        LLAMA 8B  & {0.2366} & {0.3038} & {0.1011} & {0.1584} \\
        GEMMA 9B  & \textbf{0.2542} & {0.3041} & {0.1285} & {0.1407} \\
    \hline
    \hline
    \end{tabular}
        \caption{Experiment 1 results, showing the BODEGA score of TREPAT with various LLMs, averaged over all victims trained for each task.}
    \label{tab:exp1}
\end{table}
\begin{table}
    \small
    \begin{tabular}{r|rrrr}
    \hline
    & \multicolumn{4}{|c}{\textbf{BODEGA score}}\\
    \textbf{Prompt} & \textbf{PR} & \textbf{FC} & \textbf{RD} & \textbf{HN} \\
    \hline
        REPHRASE  & {0.2420} & {0.3035} & {0.1313} & {0.1420} \\
        PARAPHRASE  & {0.2361} & {0.3027} & {0.1221} & {0.1466} \\
        SIMPLIFY  & {0.2400} & {0.2909} & {0.1344} & {0.1567} \\
        FORMAL  & {0.2400} & {0.2939} & \textbf{0.1493} & {0.1525} \\
        INFORMAL  & \textbf{0.2631} & \textbf{0.3242} & {0.1298} & \textbf{0.1780} \\
        CHANGE  & {0.2478} & {0.3016} & {0.1286} & {0.1453} \\
    \hline
        & \multicolumn{4}{|c}{\textbf{Semantic score}}\\
    \hline
        REPHRASE  & {0.7550} & {0.7830} & {0.8629} & {0.9443} \\
        PARAPHRASE  & \textbf{0.7620} & {0.7930} & {0.8534} & {0.9435} \\
        SIMPLIFY  & {0.7615} & {0.7899} & \textbf{0.8687} & {0.9431} \\
        FORMAL  & {0.7568} & {0.7953} & {0.8556} & {0.9350} \\
        INFORMAL  & {0.7508} & \textbf{0.8032} & {0.8611} & {0.9425} \\
        CHANGE  & {0.7618} & {0.7915} & {0.8676} & \textbf{0.9456} \\
    \hline
    \hline
    \end{tabular}
        \caption{Experiment 2 results, showing the BODEGA and semantic score of OLMO-based TREPAT with various prompts, averaged over victims for each task.}
    \label{tab:exp2}
\end{table}
\textbf{Experiment 1: }Table \ref{tab:exp1} shows the result of the LLM selection. We can see that there is no one model that dominates across the board. Instead, in each task a different LLM achieves the best score and the differences between them are quite limited. Therefore, we have decided to use OLMO due to its open and transparent features \citep{groeneveld-etal-2024-olmo}, as opposed to the commercial models.

\begin{table*}
    \small
    \centering
    \begin{tabular}{rrrrrrr}
    \hline
    \textbf{Task} & \textbf{Prompt} & \textbf{BODEGA} & \textbf{Confusion} & \textbf{Semantic} & \textbf{Character} & \textbf{Queries}   \\
    \hline
        PR & BERT-ATTACK  & \textbf{0.2307} & {0.3462} & {0.7221} & \textbf{0.9186} & {40.4146} \\
         & F-BERT-ATTACK  & {0.2260} & {0.3462} & {0.7095} & {0.9154} & {36.1707} \\
         & BeamAttack  & {0.1711} & {0.2404} & \textbf{0.7832} & {0.8946} & {46.1220} \\
         & \hl{TREPAT-simple}  & {0.1560} & \textbf{0.5625} & {0.5367} & {0.4620} & {25.8822} \\
         & TREPAT-full  & \textbf{0.2307} & {0.3870} & {0.7124} & {0.8159} & {27.9279} \\
    \hline
        FC & BERT-ATTACK  & {0.2289} & {0.3086} & {0.7649} & \textbf{0.9693} & {46.6148} \\
         & F-BERT-ATTACK  & {0.1216} & {0.1679} & {0.7520} & {0.9622} & {45.4988} \\
         & BeamAttack  & {0.0876} & {0.1012} & \textbf{0.8982} & {0.9614} & {49.4938} \\
         & \hl{TREPAT-simple}  & \textbf{0.3783} & \textbf{0.6444} & {0.7317} & {0.7785} & {25.3358} \\
         & TREPAT-full  & {0.3348} & {0.4444} & {0.8175} & {0.9167} & {33.5605} \\
    \hline
        RD & BERT-ATTACK  & {0.0271} & {0.0530} & {0.5256} & \textbf{0.9727} & {48.9952} \\
         & F-BERT-ATTACK  & {0.0292} & {0.0627} & {0.4821} & {0.9705} & {47.9229} \\
         & BeamAttack  & {0.0308} & {0.0361} & \textbf{0.8842} & {0.9635} & {49.5942} \\
         & \hl{TREPAT-simple}  & {0.0987} & \textbf{0.1711} & {0.6949} & {0.7450} & {44.5253} \\
         & TREPAT-full  & \textbf{0.1176} & {0.1422} & {0.8696} & {0.9411} & {45.3590} \\
    \hline
        HN & BERT-ATTACK  & {0.0000} & {0.0000} & {0.0000} & {0.0000} & {50.0000} \\
         & F-BERT-ATTACK  & {0.0732} & {0.1100} & {0.6691} & \textbf{0.9939} & {46.2150} \\
         & BeamAttack  & {0.0000} & {0.0000} & {0.0000} & {0.0000} & {50.0000} \\
         & \hl{TREPAT-simple}  & \textbf{0.2646} & \textbf{0.3000} & {0.9012} & {0.9772} & {38.1675} \\
         & TREPAT-full  & {0.1719} & {0.1850} & \textbf{0.9362} & {0.9920} & {44.1525} \\    \hline
    \hline
    \end{tabular}
        \caption{Evaluation results showing the performance of \hl{TREPAT variants} and baselines, applied to \textbf{BERT} victim models trained for the four tasks (results for BiLSTM, GEMMA2B and GEMMA7B victims are available in the appendix). For each run, the mean BODEGA, confusion, semantic and character scores are included, as well as the number of queries.}
    \label{tab:expbert}
\end{table*}

\textbf{Experiment 2:} Table \ref{tab:exp2} includes the results, with the upper half showing the BODEGA score achieved with the TREPAT method using various prompt types. Interestingly, the best performance is achieved by prompts that perform style transfer towards a style that differs from the original text: INFORMAL rephrasing for text from journalistic (PR, HN) or encyclopaedic (FC) sources, and FORMAL for the task with social media messages (RD). Other approaches also perform well, but not quite as the style transfer. 

Additionally, we verify whether the successful prompts do not harm the meaning preservation by checking the semantic score (bottom half of Table \ref{tab:exp2}). They all seem quite similar in that respect, with the differences mostly within 1\% range. For the final evaluation we choose the FORMAL (for RD) and INFORMAL (for PR, FC and HN) prompts.

\textbf{Experiment 3:} Table \ref{tab:expbert} shows the detailed attack summary for the BERT victim models. The results for the other victims (BiLSTM, GEMMA2B and GEMMA7B) are shown in Appendix \ref{sec:otherresults} (Tables \ref{tab:expbilstm}, \ref{tab:expgemma} and \ref{tab:expgemma7b}) and paint a broadly similar picture.

We see that in propaganda recognition task, the simplest solution (BERT-ATTACK) works the best, or equally well as TREPAT-full in case of BERT victims. This task involves very short fragments (average length of 24.4 words), which means the 50 queries are sufficient to find an AE for a substantial number of cases (confusion score above 30\%). 

In case of fact-checking, the fragments are slightly longer (average of 41.3 words), since they include both a claim and evidence necessary to verify it. In this situation, both TREPAT variants achieve superior results. This involves finding AEs for more examples, even if not all of them have the highest semantic similarity. This aspect is better addressed by BeamSearch, but the lower number of successes (e.g. 10\% in BERT-FC, compared to 64\% of TREPAT-simple) limits the overall score.

The RD and HN tasks both include very long text fragments: Twitter threads (average of 320.4 words) and news articles (average of 708.6 words). BERT-ATTACK and BeamAttack are clearly constrained by the victim usage limit: the number of queries asked gets close to or reaches the limit of 50. For hyperpartisan news, this situation happens for every instance, resulting in a BODEGA score of 0.0. TREPAT-simple thrives in these conditions, delivering the best overall score in both tasks.

Taking into account all the victims (see Appendix \ref{sec:otherresults}), the TREPAT variants obtain the highest BODEGA score for 15 out of 16 tested scenarios. The text length plays a role, with our approach dominating BERT-ATTACK for longer examples. The F-BERT-ATTACK variant has non-zero scores in more scenarios, but it achieves only one top spot. The BeamAttack's limited scope is reflected by confusion score, but where it is successful, the AEs have very high semantic similarity to the original text. The victims based on modern large LLMs are not necessarily more robust than BERT, aligning with the observations for simpler AE generators in the same task \citep{verifying_robustness}.

\hl{Regarding the TREPAT variants, we can see that each has its strengths. The simple version performs more aggressive rephrasing, reaching higher confusion rates with less queries, obtaining the best result for FC and HN. The full version gradually applies small changes, which requires more queries, but guarantees better semantic similarity, with the best BODEGA score for PR and RD.}

\begin{figure}
\centering
\includegraphics[width=\linewidth]{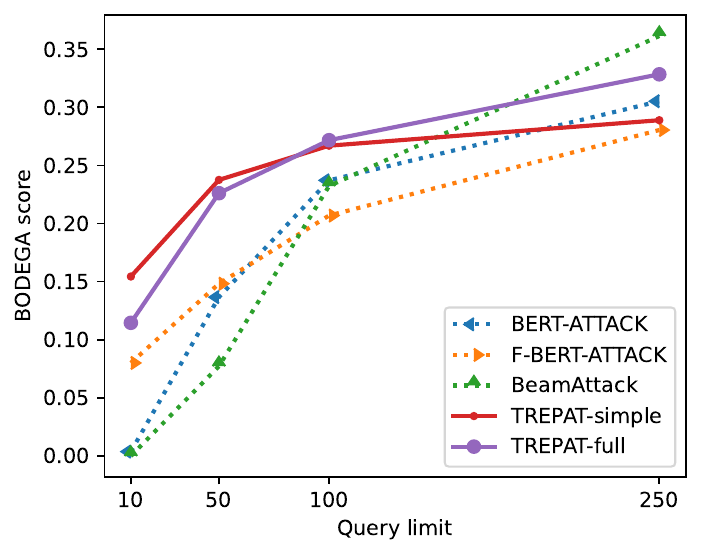} 
\caption{Experiment 4 results, showing the BODEGA score (averaged over victims and tasks) of various methods, evaluated with a given victim query limit (x axis).}
\label{fig:limits}
\end{figure}

\textbf{Experiment 4:} Figure \ref{fig:limits} shows the performance of the tested methods for various limits of queries allowed for each example. We can see that TREPAT in both variants clearly outperforms baselines within 10-100 range reported as typical daily limits in social media sites. We need to allow 250 queries to see the advantage of methods designed for unlimited queries (BeamAttack).

\subsection{Manual evaluation}
\label{sec:manualresults}

The data prepared for manual evaluation according to the procedure in Section \ref{sec:manualevaluation} included 350 instances: 165 from PR, 100 from FC, 41 from RD and 44 from HN. Of the 16 task/victim combinations, BERT-ATTACK was used as a baseline in 7, F-BERT-ATTACK in 8 and BeamAttack in 1.

We calculated the inter-annotator agreement by taking the cases where both annotators made a clear decision (A or B) and computing in how many of these they agree. The result is 75\% for meaning preservation and 63\% for language naturalness, reflecting the subjective nature of the task.

\begin{table}
    \small
    \begin{tabular}{r|rrrr}
    \hline
    & \multicolumn{4}{|c}{{Meaning preservation}}\\
    \textbf{} & \textbf{PR} & \textbf{FC} & \textbf{RD} & \textbf{HN} \\
    \hline
        Baseline  & {36.43\%} & {34.18\%} & {30.56\%} & {25.00\%} \\
        TREPAT  & \textbf{63.57\%} & \textbf{65.82\%} & \textbf{69.44\%} & \textbf{75.00\%} \\
    \hline
        & \multicolumn{4}{|c}{{Language naturalness}}\\
    \hline
        Baseline  & {47.48\%} & \textbf{59.52\%} & {48.48\%} & {40.63\%} \\
        TREPAT  & \textbf{52.52\%} & {40.48\%} & \textbf{51.52\%} & \textbf{59.38\%} \\
    \hline
    \hline
    \end{tabular}
        \caption{Results of the manual evaluation in each task, expressed as a percentage of cases where either TREPAT or the baseline approach were chosen as preferred.}
    \label{tab:annotation}
\end{table}

Table \ref{tab:annotation} shows the results of the manual annotation, computed based on cases where either one or both annotators preferred one of the options. We can see that in all tasks, the annotators judged the changes proposed by TREPAT as better at preserving meaning. With the exception of fact-checking, TREPAT also offered more natural language, confirming the validity of the attacks. 

It is interesting to notice that in terms of meaning preservation, the proposed method has the biggest advantage over the baseline in the rumour and hyperpartisan news detection tasks. These are also the tasks that are the most challenging according to the automatic evaluation (see Section \ref{sec:automaticresults}), with the low confusion score of both TREPAT and the baselines expressing the difficulty of finding an AE within the limited number of queries.

In terms of language naturalness, the gains are not as strong and in some cases the baselines offer more believable language, especially for fact-checked claims. See further discussion on the failures observed in Section~\ref{sec:ling}. 

\subsection{Linguistic analysis}
\label{sec:ling}
In order to be robust in real-world scenario, AEs must also produce utterances which are grammatical and authentic
 and also believable according to the consumer's world knowledge.
  TREPAT was the superior approach in terms of meaning preservation and was favoured most of the time for naturalness, using many strategies to modify the base text.

In this section, we describe frequent strategies that the attackers use to adhere to the heuristics set out for manual evaluation along with exemplars which can be found in Appendix~\ref{sec:quali}. Tokens and characters which have been changed between the original and modified texts are in \textbf{bold} text.

For an example of a successful modification, consider example (1) in Appendix~\ref{sec:quali} from the PR task: we observe that TREPAT uses a familiar rephrasing ``rowed back''$\rightarrow$``it revised'' and semantically bleaches the noun phrase ``verified facts'' to ``confirmed data''. This does not alter the text meaning.

Compare it to BERT-ATTACK, which changes only one word, the proper noun ``Guardian'' to ``forward''. Unlike the TREPAT rephrasing, this seemingly light-touch approach is unsuccessful and jarring to a reader by not only making the phrase ungrammatical - but also removing a key piece of information from the phrase (a newspaper name). This strategy has been observed in other AE studies \cite{przybyla-etal-2024-know} for this model.

TREPAT rephrasing is based on LLMs prone to hallucination, however. Examples (2a) and (2b) from FC shows that this can affect naturalness, with a noun phrase which appears misplaced or unnecessarily repeated. For example, TREPAT often introduces repetition of information (Example (3)) or individual words -- such as ``*their sincerely-held \textbf{belief} beliefs'' from HN -- in its paraphrasing. As shown in Table~\ref{tab:annotation}, TREPAT retains meaning better but is considered less natural for the FC task.

Two of the modification tasks described in Section~\ref{sec:rephrasing} -- introducing a more formal or informal style -- are mostly successful (Example (4)), especially in retaining meaning and naturalness with a couple of exceptions (Example (5)).

TREPAT AEs appear to be less liable to violate grammatical rules like verbal agreement compared to other models. For example, BERT-ATTACK in Example (4) ``*they know\textbf{s}'' or F-BERT-ATTACK in the RD task ``*make them suffer\textbf{ing} a trial''. Many generate highly idiomatic structures and phrases such as ``right in the gut'' in the PR task. 

\section{Conclusions}

We have presented a method to harness the text generation potential of large language model and apply it to the task of generating adversarial examples. It can to attack misinformation detection classifiers, while maintaining realistic limits on the interaction with the victim and preserving the meaning of the original example. \hl{While the use of LLM for adversarial examples has been explored before, it has not been successful} \citep{clef-checkthat:2024:turquaz} \hl{. Our work is the first to show that this method can improve the results, establishing SOTA on the BODEGA task. We are also the first to propose splitting rephrasings into edit operations to preserve semantic content.} Interestingly, the large modern models are not necessarily more robust to our attacks, emphasising the need to analyse the vulnerability of any ML solutions deployed in such a sensitive role as content moderation in social media. 

\section*{Acknowledgements}
The work of P.\@ Przyby{\l}a is part of the ERINIA project, which received funding from the European Union’s Horizon Europe research and innovation programme under grant agreement No 101060930.  Views and opinions expressed are however those of the author(s) only and do not necessarily reflect those of the funders. Neither the European Union nor the granting authority can be held responsible for them. We gratefully acknowledge Polish high-performance computing infrastructure PLGrid (HPC Centers: ACK Cyfronet AGH) for providing computer facilities and support within computational grant no. PLG/2025/018019. We also acknowledge support from Departament de Recerca i Universitats de la Generalitat de Catalunya (ajuts SGR-Cat 2021) and from  Maria de Maeztu Units of Excellence Programme CEX2021-001195-M, funded by MCIN/AEI /10.13039/501100011033. Finally, we are grateful for the participation of Alba Táboas García in the manual evaluation effort.

\section*{Ethical impact}

As with any research in the area of credibility and misinformation, we need to consider whether our work can be useful for malicious actors. We do try to make our attack scenario as close to real world as possible (e.g. limiting the number of queries), but several differences remain that make it impossible to use our method directly for performing attacks. \hl{The chief among them is the usage of victim's decision as a continuous number (e.g. 27\% credible) instead of a binary decision (e.g. REJECTED) that would typically be the only output seen by a user. This is a conscious decision, resulting from trying to balance the realistic setup with avoiding creating tools that can be directly used by attackers. Requiring a continuous score lets us provide a solution that can be used by the services deploying content filtering models, but not the attackers. This assumption is common in AE generation field, including the framework we use for evaluation.} 

Moreover, we need to note that the very practice of using automatic ML-based solutions for content filtering is considered unethical by many, e.g. because of equivalence with censorship according to the international law \citep{Llanso2020}. Nevertheless, it remains widely used by platforms and may be unavoidable given the amount of content they need to scrutinise.

\section*{Limitations}

The results show that TREPAT works as expected, delivering many adversarial examples in limited query scenarios, even when dealing with very long text. However, some limitations remain.

Firstly, while we have modified BERT-ATTACK to make it a better fit for the constrained queries scenario, no equally obvious modification was performed for BeamAttack. Employing the reduced parameter setting was clearly not enough and we see the method attempting to send too many queries. We expect this and other methods could be tuned to deliver better AEs in this setting, but this is left for future work.

Moreover, a manual analysis of the results points to an important limitation of LLMs compared to simpler models: they often avoid generating text that might be considered sensitive, toxic or crude, resorting to euphemistic replacements. Unfortunately, such topics are prevalent in discussions adjacent to misinformation. The indirect paraphrases proposed by LLMs do not fit the context style and stray too far from the original to be useful for the task. \hl{Most of these are then removed through the filtering mechanism in TREPAT (section }\ref{sec:changes}\hl{), but future work may lead to better solutions.} This relates to a wider topic of LLM \textit{exaggerated safety} \cite{chehbouni-etal-2024-representational}. 

It is important to note that while here we argue that the limited-query scenario is close to the real-world situation, other attack scenarios may be applicable. For example, one in which an attacker uses a history of previous attacks on the same victim to be able to deliver more precise AEs later. Situations when unlimited queries are allowed can also happen, e.g. if a filtering system is open enough for an attacker to deploy a local copy for their use. This only emphasises that every time a text classifier is used in adversarial scenario, it needs to be first tested for robustness against attacks that are possible in this particular application.

\hl{Finally, we need to note that while the simulated nature of the AE search performed here and in other works in the domain remains its strong limitation, there is no clear alternative. One might imagine an attempt to attack real-world systems in order to test their robustness against certain prohibited content types (e.g. misinformation), but that could cause significant harm. Such actions would be illegal in most countries, breaking the terms and conditions of the services, and risking introducing further misinformation into the media, if successful. Therefore, attempting to simulate attacks under largely realistic conditions remains the best way to make services robust against adversarial actors.}
\bibliography{trepat}

\appendix

\section{Prompts}
\label{sec:prompts}

The following prompts were used to obtain rephrasings of a given fragment:
{
\setlength{\parskip}{-0.5em}
\begin{itemize}
\setlength\itemsep{-0.2em}
    \item \texttt{REPHRASE}: \textit{\textbf{Rephrase} the provided input text. You can add, remove or replace individual words or punctuation characters, but \textbf{keep the changes to the minimum to preserve the original meaning}. Return five different rephrasings, separated by newline. Do not generate any text except the reformulations.\\INPUT:\\<fragment>\\OUTPUT:}
    \item \texttt{PARAPHRASE}: \textit{\textbf{Paraphrase} the provided input text. You can add, \ldots} (continued as above)
    \item \texttt{SIMPLIFY}: \textit{\textbf{Simplify} the provided input text. You can add, \ldots} 
    \item \texttt{FORMAL}: \textit{Rewrite the provided input text in a \textbf{more formal style}. You can add, \ldots}
    \item \texttt{INFORMAL}: \textit{Rewrite the provided input text in a \textbf{less formal style}. You can add, \ldots}
    \item \texttt{CHANGE}: \textit{\textbf{Make changes} to the provided input text. You can add, remove or replace individual words or punctuation characters, but \textbf{try to preserve the original meaning}. Return \ldots} 
\end{itemize}
}

\section{Examples for qualitative analysis}
\label{sec:quali}

Example (1): \textit{PR2.193} TREPAT does ``light touch'' edits more successfully.

Original text: faced with the backlash , the \textbf{Guardian} quickly – and very quietly – rowed back its initial certainty that its story was based on verified facts.

TREPAT: faced with the backlash , the guardian quickly – and very quietly – \textbf{it revised} its initial certainty that its story was based on confirmed data.

BERTattack: faced with the backlash, the \textbf{forward} quickly – and very quietly – rowed back its initial certainty that its story was based on verified facts.

\vspace{0.4cm}

\noindent Example (2a): \textit{FC.184} TREPAT loses naturalness but violates meaning preservation less seriously.

Original text: the disaster artist ( \textbf{film} ). it had its world premiere at south by southwest on march 12, 2017. 
$\rightarrow$ the disaster artist ( film ) is \textbf{8} years old.

TREPAT: the disaster artist \textbf{movie debut} : it had its world premiere at south by southwest on march 12, 2017. 
$\rightarrow$ the disaster artist ( film ) is 8 years old.

BERTattack: the disaster artist ( film ). it had its world premiere at south by southwest on march 12, 2017. 
$\rightarrow$ the disaster artist ( film ) is \textbf{1} years old.

\vspace{0.4cm}

\noindent Example (2b): \textit{FC.0} TREPAT repeats information unnaturally but retains meaning preservation.

Original text: $\rightarrow$ hannah and her sisters is an american \textbf{1986} \textbf{film}

TREPAT: $\rightarrow$ hannah and her sisters is an american 1986 \textbf{american movie}

BERTattack: $\rightarrow$ hannah and her sisters is an american \textbf{1987} film

\vspace{0.4cm}

\noindent Example (3): \textit{FC.128} TREPAT repeating entire phrases with the same information phrased slightly differently.

Original: colin kaepernick. kaepernick was selected by the \textbf{san francisco 49ers in the second round of} the 2011 nfl draft. 
$\rightarrow$ colin kaepernick starts for the \textbf{san} francisco 49ers.

TREPAT: colin kaepernick. kaepernick was selected by the \textbf{2011 nfl draft, kaepernick was selected by} the 2011 nfl draft. 
$\rightarrow$ colin kaepernick starts for the san francisco 49ers.

BERTattack: colin kaepernick. kaepernick was selected by the san francisco 49ers in the second round of the 2011 nfl draft. 
$\rightarrow$ colin kaepernick starts for the \textbf{special} francisco 49ers.

\vspace{0.4cm}

\noindent Example (4): \textit{PR2.42} TREPAT reduces formality but sentence is still acceptable.

Original: we \textbf{failed to recall} that \textbf{they know nothing at all} about catholicism.

TREPAT: we \textbf{forgot} that they \textbf{don't have a clue} about catholicism.

BERTattack: we failed to recall that they \textbf{knows} nothing at all about catholicism.

\vspace{0.4cm}

\noindent Example (5): \textit{PR2.115} TREPAT uses informality to create a bizarre tone, but maintaining the correct number for meaning preservation.

Original: sleipnir has \textbf{eight legs.}

TREPAT: sleipnir has eight \textbf{hooves!}

BERTattack: sleipnir has \textbf{six} legs.

\vspace{0.4cm}

\section{Annotation guidelines}
\label{sec:guidelines}

\begin{figure*}
\centering
\includegraphics[width=\linewidth]{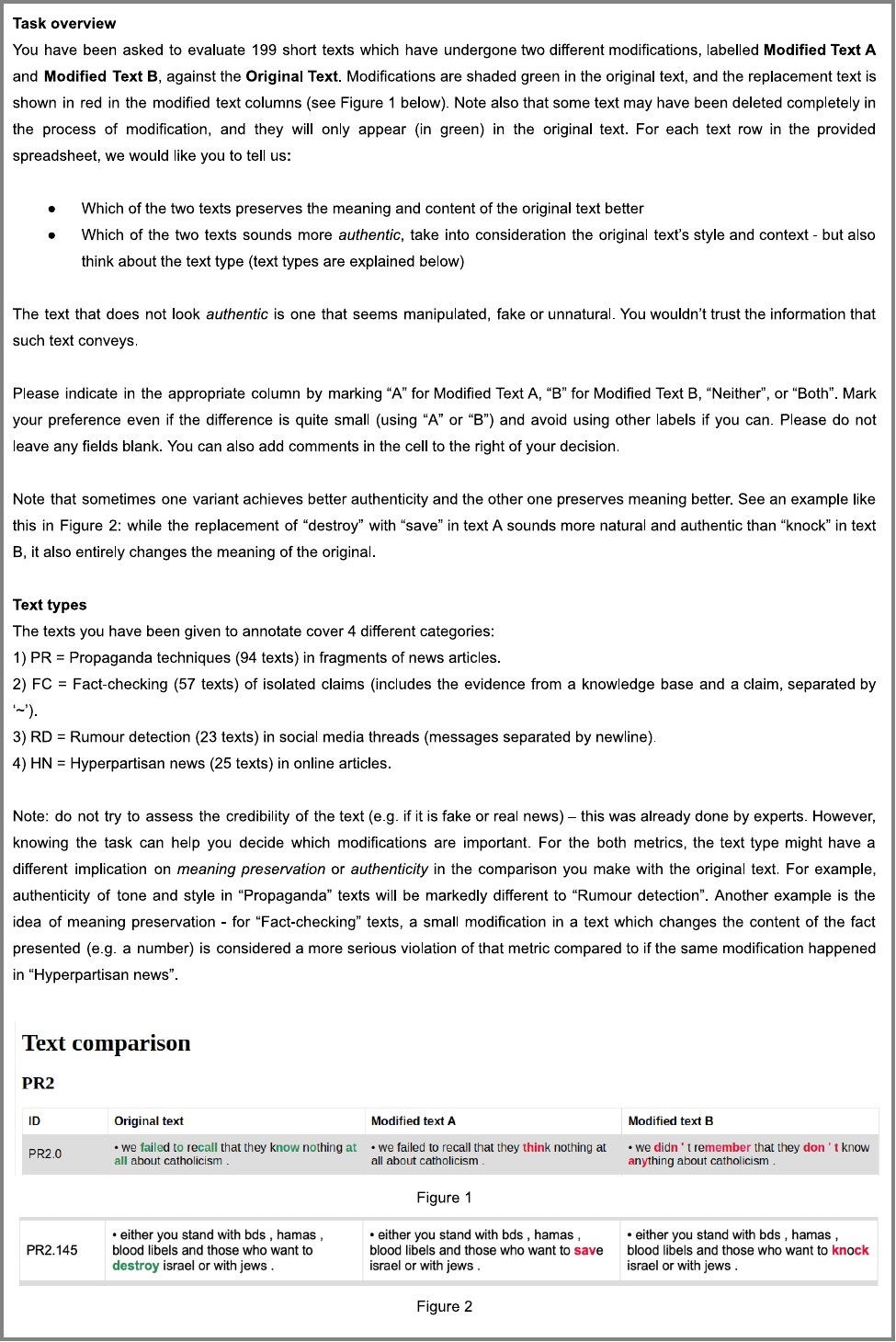} 
\caption{Annotation guidelines provided to the annotators.}
\label{fig:guidelines}
\end{figure*}

Figure \ref{fig:guidelines} presents the annotation guidelines provided to the linguists performing the manual evaluation of meaning preservation and language naturalness.

\section{Results for other victims}
\label{sec:otherresults}

Tables \ref{tab:expbilstm}, \ref{tab:expgemma} and \ref{tab:expgemma7b} show the full results of Experiment 3 for the BiLSTM, GEMMA2B and GEMMA7B victims, respectively.

\begin{table*}
    \small
    \centering
    \begin{tabular}{rrrrrrr}
    \hline
    \textbf{Task} & \textbf{Prompt} & \textbf{BODEGA} & \textbf{Confusion} & \textbf{Semantic} & \textbf{Character} & \textbf{Queries}   \\
    \hline
        PR & BERT-ATTACK  & {0.3312} & {0.4760} & {0.7462} & \textbf{0.9287} & {37.0439} \\
         & F-BERT-ATTACK  & {0.3001} & {0.4543} & {0.7198} & {0.9123} & {32.7260} \\
         & BeamAttack  & {0.2382} & {0.3413} & \textbf{0.7763} & {0.8812} & {45.0463} \\
         & 
         \hl{TREPAT-simple}  & {0.2174} & \textbf{0.7548} & {0.5568} & {0.4704} & {18.5024} \\
         & TREPAT-full  & \textbf{0.3493} & {0.5457} & {0.7473} & {0.8399} & {21.8462} \\
    \hline
        FC & BERT-ATTACK  & {0.2498} & {0.3333} & {0.7696} & \textbf{0.9736} & {45.8222} \\
         & F-BERT-ATTACK  & {0.1798} & {0.2420} & {0.7657} & {0.9695} & {41.9778} \\
         & BeamAttack  & {0.1054} & {0.1235} & \textbf{0.8861} & {0.9620} & {49.1580} \\
         & \hl{TREPAT-simple}  & \textbf{0.5176} & \textbf{0.7901} & {0.7749} & {0.8297} & {17.5975} \\
         & TREPAT-full  & {0.4536} & {0.5753} & {0.8400} & {0.9352} & {27.3580} \\
    \hline
        RD & BERT-ATTACK  & {0.0361} & {0.0627} & {0.5919} & \textbf{0.9722} & {48.8213} \\
         & F-BERT-ATTACK  & {0.0453} & {0.0892} & {0.5242} & {0.9718} & {47.0988} \\
         & BeamAttack  & {0.0261} & {0.0337} & {0.8189} & {0.9323} & {49.6643} \\
         & \hl{TREPAT-simple}  & \textbf{0.1840} & \textbf{0.2892} & {0.7463} & {0.7967} & {41.2747} \\
         & TREPAT-full  & {0.1750} & {0.2145} & \textbf{0.8597} & {0.9396} & {42.7783} \\
    \hline
        HN & BERT-ATTACK  & {0.0000} & {0.0000} & {0.0000} & {0.0000} & {50.0000} \\
         & F-BERT-ATTACK  & {0.0998} & {0.1525} & {0.6577} & \textbf{0.9951} & {44.0250} \\
         & BeamAttack  & {0.0000} & {0.0000} & {0.0000} & {0.0000} & {50.0000} \\
         & \hl{TREPAT-simple}  & \textbf{0.3856} & \textbf{0.4325} & {0.9097} & {0.9795} & {33.5350} \\
         & TREPAT-full  & {0.1941} & {0.2075} & \textbf{0.9399} & {0.9948} & {43.5500} \\
    \hline
    \hline
    \end{tabular}
        \caption{Final evaluation results, showing the performance of \hl{TREPAT variants} and baselines, applied to \textbf{BiLSTM} victim models trained for the four tasks. For each run, the mean BODEGA, confusion, semantic and character scores are included, as well as the number of queries.}
    \label{tab:expbilstm}
\end{table*}

\begin{table*}
    \small
    \centering
    \begin{tabular}{rrrrrrr}
    \hline
    \textbf{Task} & \textbf{Prompt} & \textbf{BODEGA} & \textbf{Confusion} & \textbf{Semantic} & \textbf{Character} & \textbf{Queries}   \\
    \hline
        PR & BERT-ATTACK  & {0.2807} & {0.4087} & {0.7396} & \textbf{0.9239} & {38.6463} \\
         & F-BERT-ATTACK  & {0.2969} & {0.4399} & {0.7286} & {0.9218} & {31.7260} \\
         & BeamAttack  & {0.1627} & {0.2236} & \textbf{0.7928} & {0.8884} & {46.3317} \\
         & \hl{TREPAT-simple}  & {0.1640} & \textbf{0.7332} & {0.4420} & {0.3712} & {22.2740} \\
         & TREPAT-full  & \textbf{0.3062} & {0.4928} & {0.7283} & {0.8304} & {24.3389} \\
    \hline
        FC & BERT-ATTACK  & {0.2352} & {0.3185} & {0.7608} & {0.9701} & {46.3877} \\
         & F-BERT-ATTACK  & {0.1609} & {0.2247} & {0.7432} & {0.9623} & {43.8049} \\
         & BeamAttack  & {0.0927} & {0.1037} & \textbf{0.9160} & \textbf{0.9743} & {49.4864} \\
         & \hl{TREPAT-simple}  & \textbf{0.2998} & \textbf{0.5654} & {0.6858} & {0.7451} & {29.4790} \\
         & TREPAT-full  & {0.2392} & {0.3333} & {0.7866} & {0.9075} & {37.8074} \\
    \hline
        RD & BERT-ATTACK  & {0.0366} & {0.0699} & {0.5344} & {0.9784} & {48.7904} \\
         & F-BERT-ATTACK  & {0.0756} & {0.1614} & {0.4778} & \textbf{0.9815} & {44.6940} \\
         & BeamAttack  & {0.0274} & {0.0337} & {0.8419} & {0.9604} & {49.7133} \\
         & \hl{TREPAT-simple}  & {0.1391} & \textbf{0.2169} & {0.7427} & {0.7896} & {42.3663} \\
         & TREPAT-full  & \textbf{0.1484} & {0.1783} & \textbf{0.8626} & {0.9531} & {44.0675} \\
    \hline
        HN & BERT-ATTACK  & {0.0000} & {0.0000} & {0.0000} & {0.0000} & {50.0000} \\
         & F-BERT-ATTACK  & {0.1855} & \textbf{0.2775} & {0.6708} & \textbf{0.9964} & {37.4625} \\
         & BeamAttack  & {0.0000} & {0.0000} & {0.0000} & {0.0000} & {50.0000} \\
         & \hl{TREPAT-simple}  & \textbf{0.2016} & {0.2250} & {0.9119} & {0.9825} & {41.3275} \\
         & TREPAT-full  & {0.1410} & {0.1500} & \textbf{0.9436} & {0.9958} & {45.0275} \\
    \hline
    \hline
    \end{tabular}
        \caption{Final evaluation results, showing the performance of \hl{TREPAT variants} and baselines, applied to \textbf{GEMMA2B} victim models trained for the four tasks. For each run, the mean BODEGA, confusion, semantic and character scores are included, as well as the number of queries.}
    \label{tab:expgemma}
\end{table*}

\begin{table*}
    \small
    \centering
    \begin{tabular}{rrrrrrr}
    \hline
    \textbf{Task} & \textbf{Prompt} & \textbf{BODEGA} & \textbf{Confusion} & \textbf{Semantic} & \textbf{Character} & \textbf{Queries}   \\
    \hline
        PR & BERT-ATTACK  & \textbf{0.2502} & {0.3630} & {0.7386} & \textbf{0.9305} & {39.4878} \\
         & F-BERT-ATTACK  & {0.2453} & {0.3630} & {0.7298} & {0.9228} & {35.2668} \\
         & BeamAttack  & {0.1601} & {0.2163} & \textbf{0.8080} & {0.9058} & {45.9976} \\
         & \hl{TREPAT-simple}  & {0.1555} & \textbf{0.7476} & {0.4140} & {0.3590} & {21.7428} \\
         & TREPAT-full  & {0.2379} & {0.3918} & {0.7217} & {0.8220} & {27.4255} \\
    \hline
        FC & BERT-ATTACK  & {0.2442} & {0.3309} & {0.7596} & {0.9712} & {46.3086} \\
         & F-BERT-ATTACK  & {0.1471} & {0.2049} & {0.7433} & {0.9652} & {43.8864} \\
         & BeamAttack  & {0.1002} & {0.1111} & \textbf{0.9227} & \textbf{0.9765} & {49.5728} \\
         & \hl{TREPAT-simple}  & \textbf{0.3000} & \textbf{0.5481} & {0.6948} & {0.7621} & {29.8642} \\
         & TREPAT-full  & {0.2487} & {0.3407} & {0.7962} & {0.9118} & {37.0370} \\
    \hline
        RD & BERT-ATTACK  & {0.0363} & {0.0723} & {0.5190} & {0.9673} & {48.7639} \\
         & F-BERT-ATTACK  & {0.0510} & {0.1133} & {0.4612} & \textbf{0.9770} & {46.0313} \\
         & BeamAttack  & {0.0340} & {0.0410} & \textbf{0.8692} & {0.9528} & {49.5157} \\
         & \hl{TREPAT-simple}  & {0.1275} & \textbf{0.2193} & {0.7102} & {0.7468} & {42.0578} \\
         & TREPAT-full  & \textbf{0.1496} & {0.1831} & {0.8589} & {0.9382} & {43.4988} \\
    \hline
        HN & BERT-ATTACK  & {0.0000} & {0.0000} & {0.0000} & {0.0000} & {50.0000} \\
         & F-BERT-ATTACK  & {0.1348} & {0.2000} & {0.6763} & \textbf{0.9968} & {40.8225} \\
         & BeamAttack  & {0.0000} & {0.0000} & {0.0000} & {0.0000} & {50.0000} \\
         & \hl{TREPAT-simple}  & \textbf{0.2104} & \textbf{0.2325} & {0.9196} & {0.9837} & {41.1775} \\
         & TREPAT-full  & {0.1199} & {0.1275} & \textbf{0.9440} & {0.9963} & {46.2150} \\
    \hline
    \hline
    \end{tabular}
        \caption{Final evaluation results, showing the performance of \hl{TREPAT variants} and baselines, applied to \textbf{GEMMA7B} victim models trained for the four tasks. For each run, the mean BODEGA, confusion, semantic and character scores are included, as well as the number of queries.}
    \label{tab:expgemma7b}
\end{table*}

\end{document}